\documentclass[11pt,a4paper]{article}
\usepackage[hyperref]{acl2021}
\usepackage{times}
\usepackage{latexsym}

\usepackage{amssymb}
\usepackage{amsmath}

\usepackage{booktabs}
\usepackage{graphicx}
\usepackage{floatrow}
\newfloatcommand{capbtabbox}{table}[][\FBwidth]
\usepackage[titlenumbered,ruled,linesnumbered]{algorithm2e}
\DontPrintSemicolon

\usepackage{xspace}
\newcommand{\taskname}{HARE\xspace}

\usepackage{microtype}

\aclfinalcopy % Uncomment this line for the final submission
\title{Hone as You Read: A Practical Type of Interactive Summarization}

\author{Tanner Bohn \\
  Western University \\ 
  London, ON, Canada \\
  \texttt{tbohn@uwo.ca} \\\And
  Charles X. Ling \\
  Western University \\
  London, ON, Canada \\
  \texttt{charles.ling@uwo.ca} \\}
\date{}

\begin{document}
\maketitle
\begin{abstract}
We present HARE, a new task where reader feedback is used to optimize document summaries for personal interest \emph{during the normal flow of reading}.
This task is related to interactive summarization, where personalized summaries are produced following a long feedback stage where users may read the same sentences many times.
However, this process severely interrupts the flow of reading, making it impractical for leisurely reading.
We propose to gather minimally-invasive feedback during the reading process to adapt to user interests and augment the document in real-time.
Building off of recent advances in unsupervised summarization evaluation, we propose a suitable metric for this task and use it to evaluate a variety of approaches.
Our approaches range from simple heuristics to preference-learning and their analysis provides insight into this important task.
Human evaluation additionally supports the practicality of HARE.
The code to reproduce this work is available at \url{https://github.com/tannerbohn/HoneAsYouRead}.
\end{abstract}

\section{Introduction}
\label{sec:intro}

% contains samples of nlp short papers: https://www.emnlp-ijcnlp2019.org/program/accepted/
%Example of short papers:
%https://www.aclweb.org/anthology/D19-1647.pdf
%https://www.aclweb.org/anthology/D19-1332.pdf
%https://www.aclweb.org/anthology/D19-1614.pdf

\begin{figure}[ht]%bp]
	\centering
	% link to editable figure: https://docs.google.com/drawings/d/122WeibgQf5S5UCXva5Kh_GuGYK3IB-K2ZyXy-6wOM-0/edit?usp=sharing
	\includegraphics[width=1\columnwidth]{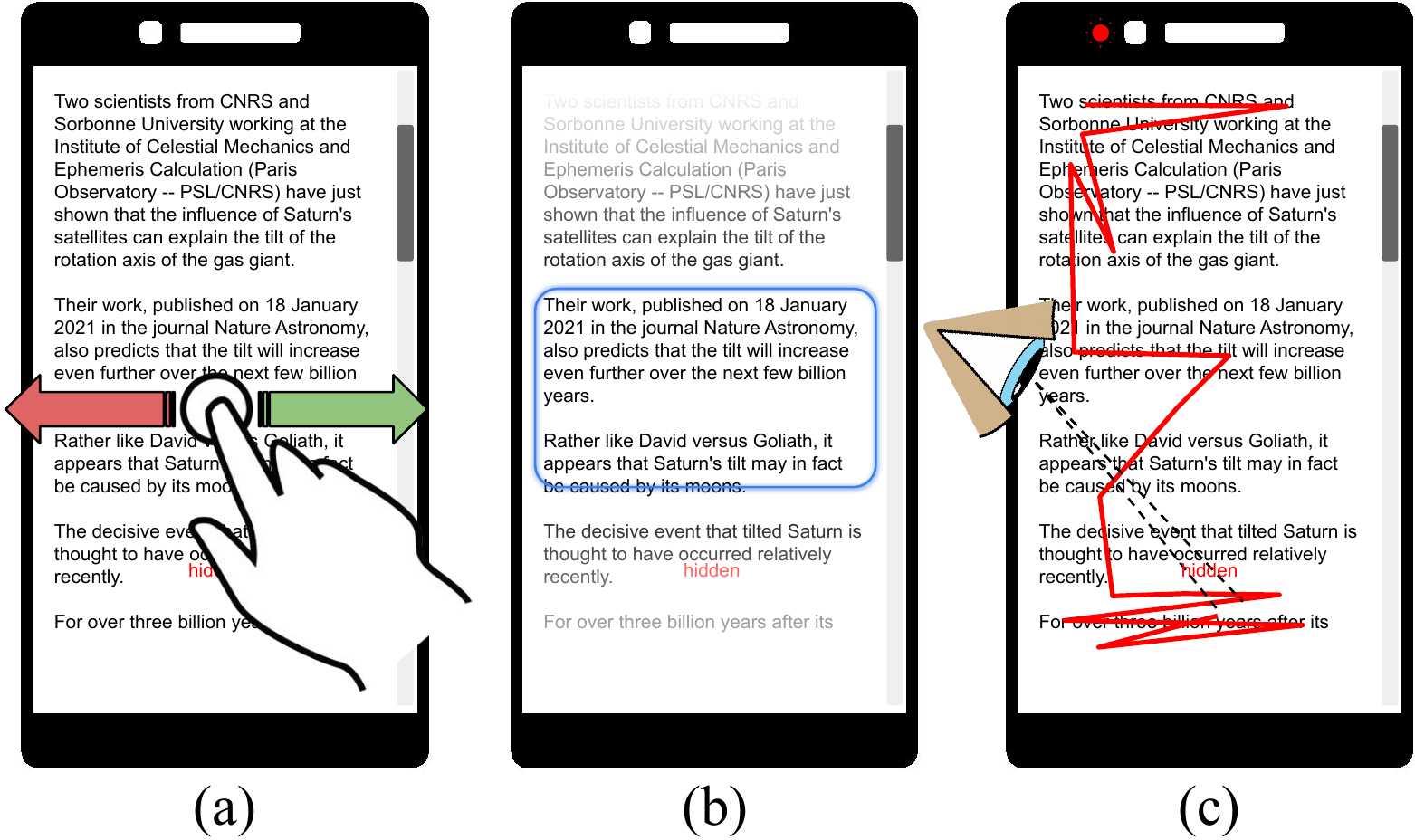}
	\caption{Potential feedback methods for \taskname used on a smartphone. In (a), users can choose to swipe left or right to indicate interest or disinterest in sections of text as they read. Users may also provide implicit feedback in the form of dwell time in center window (b) or gaze location, as measured by camera for example (c). More interesting text may have longer gazes or dwell time. The approaches evaluated in this paper rely on feedback similar to (a), but further development in \taskname can extend to (b) or (c).}
	\label{fig:feedback_alternatives}
\end{figure}

Keeping readers engaged in an article and helping them find desired information are important objectives \cite{calder2009experimental,nenkova2011automatic}. These objectives help readers deal with the explosion of online content and provide an edge to content publishers in a competitive industry.
To help readers find personally relevant content
%with minimal intrusiveness, 
while maintaining the flow of natural reading,
we propose a new text summarization problem where the summary is \textbf{h}oned \textbf{a}s you \textbf{re}ad (HARE). %maintaining the flow of natural reading. 
The challenge is to learn from unobtrusive user feedback, such as the types in Figure~\ref{fig:feedback_alternatives}, to identify uninteresting content to hop over.

% ISPS: Interactive Single-Pass Personalized Summarization
% YORO: You only read once
% HARE: Hone as you REad

This new task is related to both query-based summarization (QS) and interactive personalized summarization (IPS). In QS, users must specify a query to guide the resultant summary \cite{damova2010query}.
For users performing focused research, specifying queries is useful, but for more leisurely reading, this requirement interrupts the natural flow. 
Approaches to IPS avoid the problem of having to explicitly provide a query. However, they suffer a similar problem by requiring users to go through several iterations of summary reading and feedback-providing before a final summary is produced \cite{yan2011summarize,avinesh2018sherlock,gao2019preference,simpson2019interactive}.

In contrast, \taskname places high importance on non-intrusiveness by satisfying multiple properties detailed in Section~\ref{sec:interaction_loop} (such as feedback being non-invasive).
%requiring that 1) feedback is optional and non-invasive, 2) article sentences are shown in their original order, and 3) updates to the article (adding or removing sentences) can only happen in the portion of the article that is not yet read. 
We find that due to the high cost of generating a dataset for this task, evaluation poses a difficulty. To overcome this, we adapt recent research in unsupervised summary evaluation.
We also describe a variety of approaches for \taskname that estimate \textit{what} the user is interested in and \textit{how much} they want to read. Automated evaluation finds that relatively simple approaches based on hiding sentences nearby or similar to disliked ones, or explicitly modelling user interests, outperforms the control, where no personalization is done. 
%These techniques are able to outperform more complex methods, especially in the presence of moderate user feedback noise. TODO: how does this compare with our human experiment results?
Human evaluation suggests that not only is deciding the relevance of sentences rather easy in practice, but that even with simple binary feedback, \taskname models may truly provide useful reading assistance. 

The major contributions of this work are:
\begin{enumerate}
    \item We define the novel \taskname task, and describe a suitable evaluation technique (Section~\ref{sec:task_formulation}).
    
    \item We describe a wide range of motivated approaches for \taskname that should serve as useful baselines for future research (Section~\ref{sec:methods}).
    
    \item We evaluate our approaches to gain a deeper understanding of the task (Section~\ref{sec:experiments}).
\end{enumerate}
\section{Related Work}
\label{sec:related}

In this section, we examine related work on QS, IPS, and unsupervised summarization evaluation.

\subsection{Query-based Summarization}

Both tasks of \taskname and QS aim to produce personalized summaries.
Unlike generic summarization where many large datasets exist \cite{hermann2015teaching,alex2019multinews,Narayan2018DontGM}, development in QS has been affected by a lack of suitable training data \cite{xu2020query}. 
To cope, approaches have relied on handcrafted features \cite{conroy2005classy}, unsupervised techniques \cite{van2019query}, and cross-task knowledge transfer \cite{xu2020query}. 
%In \citet{conroy2005classy}, part-of-speech tagging and named-entity recognition is performed in order to apply a fixed set of rules in choosing summary sentences. 
The approach of \citet{mohamed2006improving} highlights how query-based summarizers often work by adapting a generic summarization algorithm and incorporating the query with an additional sentence scoring or filtering component. Alternatively, one can avoid training on QS data by decomposing the task into several steps, each performed by a module constructed for a related task \citep{xu2020query}. 
%The authors first use a graph-based representation of the text to identify central sentences, and combined it with a query-similarity factor. Such graph-based components are commonly used in generic summarization approaches \cite{mihalcea2004textrank,erkan2004lexrank}.
%
%Another unsupervised approach to QS aims to reduce redundancy by identifying ``themes'' in the text, and ensuring coverage of the query-related themes \citep{van2019query}. 
%One supervised approach is from \citet{xu2020query}. However, it avoids training on QS data by decomposing the task into several steps, each performed by a module constructed for a related task. 
%These steps include a relevance estimator, evidence estimator, and a centrality estimator.
% Each step progressively refines the final summary. The first module is a relevance estimator, which simply considers term frequency and overlap with the query. Second is the evidence estimator, which scores text spans using a model trained on a question-answering task, for which plenty of data is available. The final module is the centrality estimator, which is an extension of the LexRank summarization algorithm \cite{erkan2004lexrank}.

A pervasive assumption in QS is that users have a query for which a \emph{brief} summary is expected. This is reflected in QS datasets where dozens of documents are expected to be summarized in a maximum of 250 words \cite{dang2005overview,hoa2006overview} or single documents summarized in a single sentence \citep{hasselqvist2017querybased}.
However, in \taskname, we are interested in a wider range of reading preferences. This includes users who are interested in reading the whole article and users whose interests are not efficiently expressed in a written query.

% https://duc.nist.gov/pubs/2005papers/OVERVIEW05.pdf
%	- DUC 2005 eval only accepts up to 250 words as a summary for dozens of related documents
% https://duc.nist.gov/pubs/2006papers/duc2006.pdf
%	- DUC 2006 -- summaries are expected to be "brief"
%	- these DUC tasks are also multi-document based -- we are interested in reading of single documents
% https://arxiv.org/pdf/1712.06100.pdf
%	- adapts CNN/DM, considers each highlight sentence as a query-focused summary

%Indicative summaries can function similarly to keyphrase extraction and help the reader decide whether a text is of interest with respect to their query. Informative summaries,  

\subsection{Interactive Personalized Summarization}
% TODO:
% also similar to this: https://www.aclweb.org/anthology/N10-1041.pdf

The iterative refinement of summaries based on user feedback is also considered by IPS approaches.
An early approach by \citet{yan2011summarize} considers progressively learning user interests by providing a summary (of user-specified length) and allowing them to click on sentences they want to know more about. Based on the words in clicked sentences, a new summary can be generated and the process repeated.
Instead of per-sentence feedback, \citet{avinesh2017joint} allows users to indicate which bigrams of a candidate summary are relevant to their interests. 
A successor to this system reduces the computation time to produce each summary down to an interactive level of 500ms \cite{avinesh2018sherlock}.
% TODO: discuss the speed of our models? add a subsection to results on interactiveness?
%This is done by heuristically prioritizing candidate sentences for the next iteration of the summary. 
%
The APRIL system \cite{gao2019preference} aims to reduce the cognitive burden of IPS by instead allowing users to indicate preference between candidate summaries. Using this preference information, a summary-ranking model is trained and used to select the next pair of candidate summaries.
%
% TODO: uncomment this if there's room
%While APRIL presents summary pairs to the user for which their ranking is most uncertain, \citet{simpson2019interactive} suggests that preference information can be used more efficiently by focusing on learning to rank good candidates in particular. This insight allows their approach to significantly outperform previous ones.
%

Shared among these previous works is that the user is involved in an interactive process which interrupts the normal reading flow with the reviewing of many intermediate summaries. 
In \taskname, the user reads the document as it is being summarized, so that any given sentence is read at most once (if it has not already been removed). These previous works also focus on multi-document summarization, whereas we wish to improve the reading experience during the reading of individual documents.

\subsection{Unsupervised Summary Evaluation}

When gold-standard human-written summaries are available for a document or question-document pair, the quality of a model-produced summary is commonly computed with the ROUGE metric~\cite{lin2004automatic}.
Driven by high costs of obtaining human-written summaries at a large scale, especially for tasks such as multi-document summarization or QS, unsupervised evaluation of summaries (i.e. without using gold-standards) has rapidly developed \citep{louis2013automatically}.
%A detailed examination of the motivations of unsupervised evaluation and many proposed solutions is given by \citet{louis2013automatically}.

\citet{louis2009automatically} found that the Jensen Shannon divergence between the word distributions in a summary and reference document out-performs many other candidates and achieves a high correlation with manual summary ratings, but not quite as high as ROUGE combined with reference summaries.
\citet{sun2019feasibility} consider a variety of distributed text embeddings and propose to use the cosine similarity of summary and document ELMo embeddings \cite{peters2018deep}. 
\citet{bohm2019better} consider \emph{learning} a reward function from existing human ratings. Their reward function only requires a model summary and document as input and achieves higher correlation with human ratings than other metrics (including ROUGE which requires reference summaries). \citet{stiennon2020learning} also consider this approach, with a larger collection of human ratings and larger models.
However, \citet{gao2020supert} found that comparing ELMo embeddings or using the learned reward from \citeauthor{bohm2019better} does not generalize to other summarization tasks. Their evaluation of more advanced contextualized embeddings found that Sentence-BERT (SBERT) embeddings \cite{reimers2019sentence} with word mover's-based distance~\cite{kusner2015word} outperforms other unsupervised options. Post-publication experiments by \citeauthor{bohm2019better} further support the generalizability of this approach\footnote{The additional results can be found here: \url{https://github.com/yg211/summary-reward-no-reference}.}. 
%
%\citet{gao2020supert} also find that comparing model summaries to \textit{pseudo-reference summaries} 
%(the first several sentences of each document) 
%performs better than comparing to the entire document. 
%Given that lead sentences perform disproportionately well as summary sentence \cite{see2017get}, this is not unexpected. 
In Section~\ref{sec:metric}, we adapt the method of \citeauthor{gao2020supert} to \taskname evaluation.

\section{Task Formulation}
\label{sec:task_formulation}

To define the proposed task, we will first describe how a user interacts with an \taskname summarizer (Section~\ref{sec:interaction_loop}). 
Second, we describe a method for modelling user interests and feedback for automatic evaluation (Section~\ref{sec:user_modelling}). Third, we propose an evaluation metric for this new task (Section~\ref{sec:metric}).

\subsection{User-Summarizer Interaction Loop}
\label{sec:interaction_loop}

\begin{figure}[ht]%bp]
	\centering
	% https://docs.google.com/drawings/d/1buqC_ILa0d8GRytrzbOU1opJCQcTOwpPO_y4Wcmskh4/edit?usp=sharing
	\includegraphics[width=1\columnwidth]{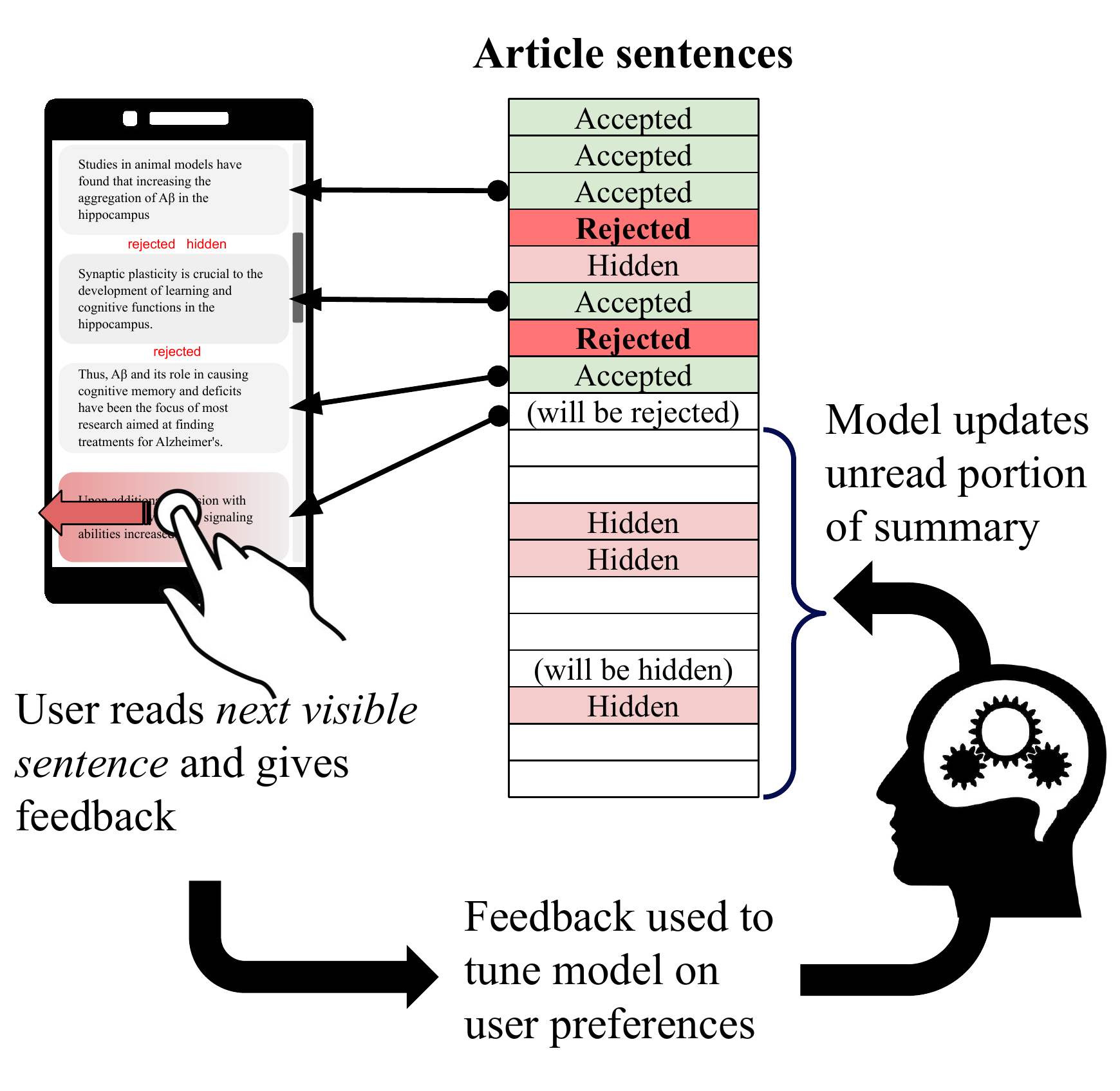}
	\caption{In \taskname, users are shown sentences in their original order, and can provide relevance feedback. A model uses this feedback to optimize the remainder of the article, automatically hiding uninteresting sentences.}
	\label{fig:demo}
\end{figure}

\begin{algorithm}[ht]
\SetAlgoLined
    user chooses a document $D = [x_{1}, ..., x_{|D|}]$ to read with help from summarizer $M$\;
    $S = \emptyset$ \tcp{summary sentences}
    \For{i = 1, ..., $|D|$}{
        \If{$M$ decides to show $x_{i}$ to user}{
            show sentence $x_{i}$ to user\;
            $S := S \cup \{x_{i}\}$\;
            incorporate any feedback into $M$\;
        }
        \If{user is done reading}{break}
    }
\Return S
\caption{User-Summarizer Interaction}
\label{alg:interaction}
\end{algorithm}

The interaction between a user and \taskname summarizer, as shown in Figure~\ref{fig:demo} and sketched in Algorithm~\ref{alg:interaction}, consists of the user reading the shown sentences and providing feedback on their relevance. Using this feedback, the summarizer decides which remaining sentences to show, aiming to hide uninteresting sentences.
This interaction is designed to smoothly integrate into the natural reading process by exhibiting three important properties: 1) feedback is either implicit or non-intrusive, 2) sentences are presented in their original order to try maintain coherence, and 3) updates to the summary should occur beyond the current reading point so as to not distract the user.
Next, we discuss how to model a user in this interaction for the purposes of automatic evaluation.

% mention how the reason past sentences cannot be changed is that it means the reader needs to interrupt their reading to go back
%While this interaction model makes several simplifying assumptions, it still poses a sufficient challenge for a first exploration of the task. To move past this simple model, future work may consider allowing the user to read the document in a non-linear fashion (i.e. ``jump around'') and allowing the user to provide feedback for sentences in arbitrary order.

\subsection{User Modelling}
\label{sec:user_modelling}

In order to model user interaction during \taskname, we need to know what kind of feedback they would provide when shown a sentence. This requires understanding how much a user would be interested in a given sentence and how feedback is provided.

\paragraph{User interests} 
For our work, user interests will be modelled as a weighted set of concept vectors from a semantic embedding space.
Given a weighted set of $k$ user interests, $U = \{<w_{1}, c_{1}>, ..., <w_{k}, c_{k}>\}$ such that $w_{i} \in [0, 1]$ and $\max(w) = 1$, and a sentence embedding, $x$, the interest level (which we also refer to as importance) is calculated with Equation~\ref{eqn:interest}. We use cosine distance for $\Delta$. Intuitively, the importance of a sentence reflects the maximum weighted similarity to any of the interests.  This method of computing importance is similar to that used by \citet{avinesh2018sherlock,wu2019neural,teevan2005personalizing}. However, we adapt it to accommodate modern distributed sentence embeddings (SBERT). 
%Additionally, this method is similar in design to the unsupervised evaluation metric we describe in Section~\ref{sec:metric}.
\begin{equation}
R(U, x) = \max_{i=1, ..., k} w_{i} (1 - \Delta(c_{i}, x))
\label{eqn:interest}
\end{equation}

\paragraph{Feedback types}
Given a sentence interest score of $r_{x} \in [0, 1]$, what feedback will be observed by the model?
If using implicit feedback like dwell time or gaze tracking, feedback could be continuously valued.
With explicit feedback, like ratings or thumbs up/down, feedback could be discrete. 
%Figure~\ref{fig:feedback_alternatives} shows three of these feedback options.
For an in-depth discussion on types of user feedback, see \citet{jayarathna2017analysis}.

In this work, we will consider an explicit feedback inspired by the ``Tinder sort'' gesture popularized by the Tinder dating app\footnote{\url{https://tinder.com/?lang=en}}, where users swipe left to indicate disinterest, and right to indicate interest. This feedback interaction has proven to be very quick and easy. Users will routinely sort through hundreds of items in a sitting \cite{david2016screened}.
To adapt this feedback method to our interactive summarization system, we can consider users to ``accept'' a sentence if they swipe right, and ``reject'' it if they swipe left (see Figure~\ref{fig:feedback_alternatives}a and Figure~\ref{fig:demo})\footnote{If we wanted to make the feedback optional, we could simply let no swipe indicate acceptance, and left swipe indicate rejection.}.

%To adapt this feedback method to our interactive summarization system, there are two clear feedback options. The first is \textbf{positive feedback}, where the user can ``swipe right'' to indicate that they like a sentence.
%The second option is \textbf{negative feedback}, where the user can ``swipe left'' to indicate that they dislike a sentence.
%To maintain both the optional aspect of feedback and to allow users to seamlessly read the entire document, we will only experiment with negative feedback. To understand why positive feedback is incompatible, consider these two scenarios: 1) the user wants to read the original whole article without interactivity, and 2) the user does not like any of the sentences, so does not swipe right. In both cases, the feedback is identical, even though the user preferences are very different. With negative feedback however, these two users are differentiated.

To model the noisy feedback a user provides, we adopt a logistic model, shown in Equation~\ref{eqn:neg_feedback}, following \citet{gao2019preference,viappiani2010optimal,simpson2019interactive}.
%Our feedback model is parameterized by a reaction threshold, $\alpha \in [0, 1]$, and a noise level, $m > 0$. Low $\alpha$ means that users are only likely to indicate disinterest for the worst sentences.
%We consider the model to receive a feedback value of $0$ if they swipe/dislike a sentence, and $1$ otherwise.
%Figure~\ref{fig:feedback_probs} plots swiping probabilities for various $m$ and $\alpha$.
%In setting $\alpha$ for feedback modelling, we tie it to the users length preference to better simulate realistic behavior. When users want to read very little for example, they dislike all but the best sentences. If a user wants to read $l$ out of $|D|$, then we set $\alpha = 1 - l/|D|$.
Our feedback model is parameterized by a decision threshold, $\alpha \in [0, 1]$, and a noise level, $m > 0$. Low $\alpha$ means that users are willing to accept sentences with lower importance.
We consider the model to receive a feedback value of $0$ if they reject a sentence, and $1$ if they accept.
%Figure~\ref{fig:feedback_probs} plots acceptance probabilities for various $m$ and $\alpha$.
In setting $\alpha$ for feedback modelling, we tie it to the users length preference to better simulate realistic behavior. When users want to read very little for example, they only accept the best sentences. If a user wants to read $l$ out of $|D|$, then we set $\alpha = 1 - l/|D|$. For user modelling, we sample $l$ uniformly from the range $[1, |D|]$.

%\begin{equation}
%P_{\alpha, m}(\text{swipe r. on } s) = \left[1 + exp\left(\frac{\alpha - r_{x}}{m}\right)\right]^{-1}
%\label{eqn:pos_feedback}
%\end{equation}

%https://www.wolframalpha.com/input/?i=plot+y+%3D+1+-+1%2F%281+%2B+e%5E%28-%280.2+-+x%29%2F0.01%29%29%2C+1+-+1%2F%281+%2B+e%5E%28-%280.2+-+x%29%2F0.1%29%29+for+x+%3D+0+to+1%2C+y+%3D+0+to+1
\begin{equation}
P_{\alpha, m}(\text{accept } x) = 1 - \left[1 + exp\left(\frac{\alpha - r_{x}}{m}\right)\right]^{-1}
\label{eqn:neg_feedback}
\end{equation}

\iffalse
\begin{figure}[ht]%bp]
	\centering
	\includegraphics[width=1\columnwidth]{figs/feedback_probs.pdf}
	\caption{A visualization of sentence accept probabilities for various combinations of $m$ (noise level) and $\alpha$ (decision threshold).}
	\label{fig:feedback_probs}
\end{figure}
\fi

\subsection{Unsupervised Evaluation}
\label{sec:metric}

% TODO: for future work, maybe check this out for alternative metrics: https://arxiv.org/pdf/2007.12626v3.pdf

% TODO: first re-read unsupervised evaluation section
%ROUGE CITE is the dominant metric in summarization-based tasks, which requires human-written summaries to compare model summaries against.
%However, for this new task, a suitably sized dataset would be prohibitively expensive to construct
%For this new task, constructing a suitably sized dataset of 

Unsupervised evaluation is tricky to do properly. You must show that it correlates well with human judgement, but also be confident that maximizing the metric does not result in garbage \cite{barratt2018note}.

As discussed in Section~\ref{sec:related}, we adapt the unsupervised summary evaluation method described by \citet{gao2020supert}. This metric computes a mover's-based distance between the SBERT embeddings of the summary and a heuristically-chosen subset of document sentences (a ``pseudo-reference'' summary). They show that it correlates well will human ratings and that using it as a reward for training a reinforcement learning-based summarizer produces state-of-the-art models.
The authors found that basing the pseudo-reference summary on the lead heuristic, which generally produces good single and multi-document summaries, worked best.
For \taskname, we can apply the analogous idea: when computing the summary score, we can use all document sentences in the pseudo-reference summary, but weight them by their importance: 
\begin{equation}
%M(D, S) = 1 - \frac{1}{\sum r} \sum_{i = 1, .. |D|} r_{i} \min_{j = 1, ... |S|} \Delta(i, j)
score(U, D, S) = 1 - \frac{1}{\sum_{x \in D} r_{x}} \sum_{x \in D} r_{x} \min_{s \in S} \Delta(x, s)
\label{eqn:metric}
\end{equation}
%$r_{x}$ refers to the interest value of sentence $x$ for the implied user interests. 
%One notable advantage of this metric over, e.g. adding up the interest scores of summary sentences, is that it 

%This metric rewards concept coverage, and penalizes cases where an important sentence ; an important document sentence that is ``far away'' from all summary sentences will detract from the score.
%This metric encourages each important concept to be well-represented in the summary by at least one sentence. If an important document sentence that is ``far away'' from all summary sentences will detract from the score.
This metric has the behavior of rewarding cases where an important sentence is highly similar to at least one summary sentence.
For this reason, coverage of the different user interests is also encouraged by this metric: since sentences drawing their importance from similarity to the same concept are going to be similar to each other, having summaries representing a variety of important concepts is better.

% Transforming Wikipedia into Augmented Data for Query-Focused Summarization
% https://arxiv.org/pdf/1911.03324.pdf

% Sentence Mover’s Similarity: Automatic Evaluation for Multi-Sentence Texts
% https://www.aclweb.org/anthology/P19-1264.pdf

% MoverScore: Text Generation Evaluating with Contextualized Embeddings and Earth Mover Distance
% https://www.aclweb.org/anthology/D19-1053.pdf
\section{Methods}
\label{sec:methods}

We consider three groups of approaches ranging in complexity:
(1) simple heuristics, 
(2) adapted generic summarizers, and
(3) preference learning.
%Note that we sometimes use the same letter, such as $k$, to represent a parameter in multiple approaches, even though they describe separate things. 

\subsection{Simple Heuristics}

This first set of approaches are as follows:

\paragraph{\textsc{ShowModulo}} This approach shows every $k^{th}$ sentence to the user. When $k=1$, this is equivalent to the control, where every sentence is shown. 
%When $k=2$, every second one is shown, etc. 
By moving through the article faster, we suspect that greater coverage is obtained, making it more likely that important concepts are represented.

\paragraph{\textsc{HideNext}} This approach shows all sentences, except for the $k$ following any rejected sentence. E.g. when $k=2$ and the user rejects a sentence, the two after it are hidden. The motivation for this model is that nearby sentences are often related, so if one is disliked, a neighbour might also be. Larger $k$ suggests a larger window of relatedness.

%\paragraph{\textsc{HideNextSimilar}} This approach aims to refine \textsc{HideNext} by hiding the next sentences only when they are sufficiently similar. Following sentences are only hidden until a sentence breaks the chain. For example, if the similarities following a rejected sentences are $[0.9, 0.7, 0.3, 0.9, ...]$ and the threshold is 0.6, then only the first two following are hidden. Similarity is measure with the cosine similarity of SBERT embeddings.

\paragraph{\textsc{HideAllSimilar}}
%While \textsc{HideNextSimilar} only hides the unbroken chain of similar sentences, this model hides all sentences similar to a rejected one. This approach makes more aggressive assumptions than the previous two models, but may in response adapt faster.
While \textsc{HideNext} hides physically nearby sentences, this model hides all sentences that are actually conceptually similar to a rejected one, where similarity is measure with cosine similarity of SBERT embeddings. 
%While this approach makes more aggressive assumptions than the previous one, it may in response adapt faster. 
We also include a compromise between hiding based on physical and conceptual similarity: \textbf{\textsc{HideNextSimilar}}. This model hides only the unbroken chain of similar sentences after a rejected one.

\subsection{Adapted Generic Summarizers}

This set of approaches make use of generic extractive summarizers.
The motivation for considering them is that even though they are independent of user interests, they are often designed to provide good coverage of an article. In this way, they may accommodate all user interests to some degree.
%, which have shown to perform well at certain QS tasks, despite ignoring the query TODO:CITE. 
For a given generic summarizer, we consider the following options:

\paragraph{\textsc{GenFixed}} This approach first uses the generic summarizer to rank the sentences, and then shows a fixed percentage of the top sentences.

\paragraph{\textsc{GenDynamic}} This approach estimates an importance threshold, $\hat{\alpha}$, of sentences the user is willing to read, and hides the less important sentences. Importance is computed by scoring the sentences with the generic summarizer and rescaling the values to $[0, 1]$.  The initial estimate is $\hat{\alpha}=0$, which means that all sentences are important enough. Each time a sentence is rejected, the new estimate is updated to be the average importance of all rejected sentences. To help avoid prematurely extreme estimates, we also incorporate $\epsilon$-greedy exploration. With probability $1 - \epsilon$, the sentence is only shown if the importance meets the threshold, otherwise it is shown anyways. A larger $\epsilon$ will help find a closer approximation of the threshold, but at the cost of showing more unimportant sentences.

\subsection{Preference Learning}

The approaches in this group use more capable adaptive algorithms to learn user preferences in terms of both preferred length and concepts:

\paragraph{\textsc{LR}}  This approach continually updates a logistic regression classifier to predict feedback given sentence embeddings. Before a classifier can be trained, all sentences are shown. We propose two variations of this approach. The first uses an $\epsilon$-greedy strategy similar to \textsc{GenDynamic}. The second uses an $\epsilon$-decreasing strategy: for a sentence at a given fraction, $frac$, of the way through the article, $\epsilon = (1 - frac)^{\beta}$, for $\beta > 0$.

\paragraph{\textsc{CoverageOpt}} This approach explicitly models user interests and length preference. 
It scores potential sentences by how much they improve coverage of the user interests.
%It scores potential sentences similar to the evaluation metric in Equation~\ref{eqn:metric}---aiming to achieve coverage of important sentences. 
However, since we do not know the user's true interests or their length preference, both are estimated as they read.

This approach prepares for each article by using K-Means clustering of sentence embeddings to identify core concepts of the article. The initial estimate of concept importances is computed with:
\begin{equation}
    \hat{C} = \left[1 + exp\left(\frac{cfsum}{\beta}\right)\right]^{-1}
\label{eq:concept_weight_estimate}
\end{equation}
We initialize the vector $cfsum$ with the same value $c \in \mathbb{R}$ for each concept. A larger $c$ means that more evidence is required before a concept is determined to be unimportant. $\beta > 0$ controls how \textit{smoothly} a concept shifts between important and unimportant (larger value means more smoothly).
To update the estimate of user interests with $feedback \in \{0, 1\}$ for sentence $x$, we update $cfsum$ with:
\begin{equation}
    cfsum \leftarrow cfsum + 2(feedback-0.5)concepts(x)
\label{eq:cfsum_update}
\end{equation}
If $feedback=0$ for example, this moves $cfsum$ away from the article concepts represented by that sentence.
The function $concepts()$ returns the relevance of each concept for the specified sentence. 

After updating $\hat{C}$, we re-compute sentence importances based on their contribution to improving concept coverage, weighted by concept importance.
Next, we update the estimated length preference, $\hat{l}_{frac}$, by averaging the importance of rejected sentences. 
The summary is updated to show sentences among the top $\hat{l}_{frac}|D|$ important sentences. If the user has rejected low and medium importance sentences, then only the most coverage-improving sentences will be shown.

% uses idea of saturation for computing importance
% https://www.wolframalpha.com/input/?i=plot+y+%3D+1%2F%281+%2B+e%5E%28-x%2F0.05%29%29+for+x+%3D+-1+to+1%2C+y+%3D+0+to+1

%\paragraph{\textsc{PolicyLearning}} While all previous approaches either require no training or undergo training for each article, this approach considers learning a policy that can be applied to an arbitrary article.
%We represent this policy as a multi-layer neural network which takes as input for each sentence the features listed in Table~\ref{tab:policy_features}, and outputs a single value between 0 and 1, which is interpreted at the probability of showing the sentence in question. By interpreting the output as a probability, the model is able to determine its own exploration/exploitation trade-off.
%To optimize the policy, we use the \textsc{REINFORCE} optimization algorithm TODO:CITE.
    
    %\item adapted query-based summarizers:  SQUAD relevance training (train model to take query embedding and predict how relevant other sentences are) -- similar to one of the components from xu2020?
%\input{tables/policy_features}

% ORACLES:
%- greedy oracle (why is performance not perfect? -- length limit)
%- importance oracle (worse than greedy oracle because it ignores coverage)

\section{Experiments}
\label{sec:experiments}

In this section, we first describe the experimental setup, and then provide an analysis of the results.

\subsection{Setup}

\paragraph{Dataset} We evaluate on the test articles from the non-anonymized CNN/DailyMail dataset \cite{hermann2015teaching}\footnote{Accessed through HuggingFace: \url{https://huggingface.co/datasets/cnn_dailymail}.}. We remove articles with less than 10 sentences so as to cluster sentences into more meaningful groups for user interest modelling.
This leaves us with 11222 articles, with an average of 34.0 sentences per article.

\paragraph{User modelling} We apply K-Means clustering to SBERT sentence embeddings for each article to identify $k=4$ cluster centers/concepts. User interests are a random weighting over these concepts, as described in Section~\ref{sec:user_modelling}. For feedback noise, we use $m=0.01$ (essentially no noise) and $m=0.1$ (intended to capture the difficulty in deciding whether a single sentence is of interest or not). $\alpha$ is chosen as described in Section~\ref{sec:user_modelling}. 

\paragraph{Metrics} Evaluation with the two noise values of $m=0.01$ and $m=0.1$ correspond to $score_{sharp}$ and $score_{noisy}$ respectively. $score_{adv}$ corresponds to the difference between $score_{noisy}$ and the control score (no personalization). Positive values indicate outperforming the control. Since the scores fall between 0 and 1, we multiply them by 100. 
%We also report the average number of swipes performed by the users per article.

\paragraph{Privileged information comparison models} We consider for comparison three oracle models and the control. \textsc{OracleGreedy} has access to the user preferences and greedily selects sentences to maximize the score, until the length limit is reached. \textsc{OracleSort} selects sentences based only on their interest level. \textsc{OracleUniform} selects sentences at random throughout the article until the length limit is reached\footnote{Readers cannot be guaranteed a uniform sampling of sentences unless their length preference is known in advance.}.

\subsection{Results}

Table~\ref{tab:overall} reports the results for each model with its best performing set of hyperparameters. While $score_{sharp}$ and $score_{noisy}$ can range from 0 to 100, the difference between the control and \textsc{OracleGreedy} is less that 5 points (reflected in $score_{adv}$). This suggests that even relatively small performance differences are important. For stochastic models (marked by a * in Table~\ref{tab:overall}), results are averaged across 3 trials and standard deviations were all found to be below 0.05.

Overall, we find that the simple heuristics provide robust performance, unaffected (and possibly helped) by noise. While the more complex \textsc{CoverageOpt} approach is able to perform best with low-noise feedback, it falls behind when noise increases. Next we discuss in more detail the results for each group of models, then comment on aspects of efficiency, and finally discuss the results of our human evaluation.

\begin{table}[]
\centering
\resizebox{\textwidth}{!}{%
\begin{tabular}{lccc}
\hline
Model & \multicolumn{1}{l}{$score_{sharp}$} & \multicolumn{1}{l}{$score_{noisy}$} & \multicolumn{1}{l}{$score_{adv}$} \\ \hline
\textsc{OracleGreedy}            & \multicolumn{2}{c}{87.04}       & 4.89          \\
\textsc{OracleSorted}            & \multicolumn{2}{c}{82.74}       & 0.58          \\
\textsc{OracleUniform}*           & \multicolumn{2}{c}{82.77}       & 0.62          \\
Control (show all)           & \multicolumn{2}{c}{82.15}       & 0.0          \\ \hline
\textsc{ShowModulo}              & \multicolumn{2}{c}{78.83}       & -3.32         \\
\textsc{HideNext}                & 82.66          & 82.66          & 0.51          \\
\textsc{HideNextSimilar}         & 82.79          & 82.86          & 0.71          \\
\textsc{HideAllSimilar}          & 83.03          & \textbf{83.09} & \textbf{0.94} \\ \hline
\textsc{GenFixed}                & \multicolumn{2}{c}{81.97}       & -0.19         \\
\textsc{GenDynamic}*              & 82.39          & 82.24          & 0.09          \\ \hline
\textsc{LR} ($\epsilon$-greedy)*     & 82.48          & 82.50          & 0.34          \\
\textsc{LR} ($\epsilon$-decreasing)* & 82.28          & 82.31          & 0.15          \\
\textsc{CoverageOpt}             & \textbf{83.11} & 82.81          & 0.65          \\ \hline
\end{tabular}%
}
\caption{A comparison of each model proposed. For parameterized models, results with the best variation are reported (for all models, we found that the same parameters performed best for both $score_{sharp}$ and $score_{noisy}$). Non-deterministic models are marked by a *. $score_{adv}$ is the difference between $score_{noisy}$ and the control score (which is independent of feedback).}
\label{tab:overall}
\end{table}

%%%%%%%%%%%%%%%%%%%%%%%%%%%%%%%%%%%%%%%%%%%%%%
%%%%%%%%%%%%%%%%%%%%%%%%%%%%%%%%%%%%%%%%%%%%%%
\subsubsection{Privileged Information Models}
\label{sec:results_privileged}

\textsc{OracleUniform} outperforms the control as well as \textsc{OracleSorted}. This may seem counter-intuitive, since \textsc{OracleUniform} has the disadvantage of not knowing true user interests. However, the strength of \textsc{OracleUniform} is that it provides uniform coverage over the whole article, weakly accommodating any interest distribution. By choosing only the most interesting sentences, \textsc{OracleSorted} runs the risk of only showing those related to the most important concept. 
If our user model simulated more focused interests, \textsc{OracleSorted} may perform better however.

It is also interesting to see how much higher \textsc{OracleGreedy} is than every other model, suggesting that there is plenty of room for improvement. The reason the oracle does not reach 100 is that the summary length is restricted by user preference.
%An extractive model thus cannot achieve perfect coverage of the document. 
If future approaches consider abstractive summarization techniques, it may be possible to move beyond this performance barrier.

%%%%%%%%%%%%%%%%%%%%%%%%%%%%%%%%%%%%%%%%%%%%%%
%%%%%%%%%%%%%%%%%%%%%%%%%%%%%%%%%%%%%%%%%%%%%%
\subsubsection{Simple Heuristics}

While we suspected that the \textsc{ShowModulo} strategy might benefit from exposing readers to more concepts faster, we found that this does not work as well as \textsc{OracleUniform}. The top performance of $score_{adv} = -3.32$ is reached with $k=2$, and it quickly drops to $-7.06$ with $k=3$.
The minimally adaptive approach of hiding a fixed number of sentences after swiped ones, as per \textsc{HideNext}, does help however, especially with $n=2$. 
%More detailed results for both of these models can be found in the appendix.

\begin{figure}[ht]%bp]
	\centering
	\includegraphics[width=1\columnwidth]{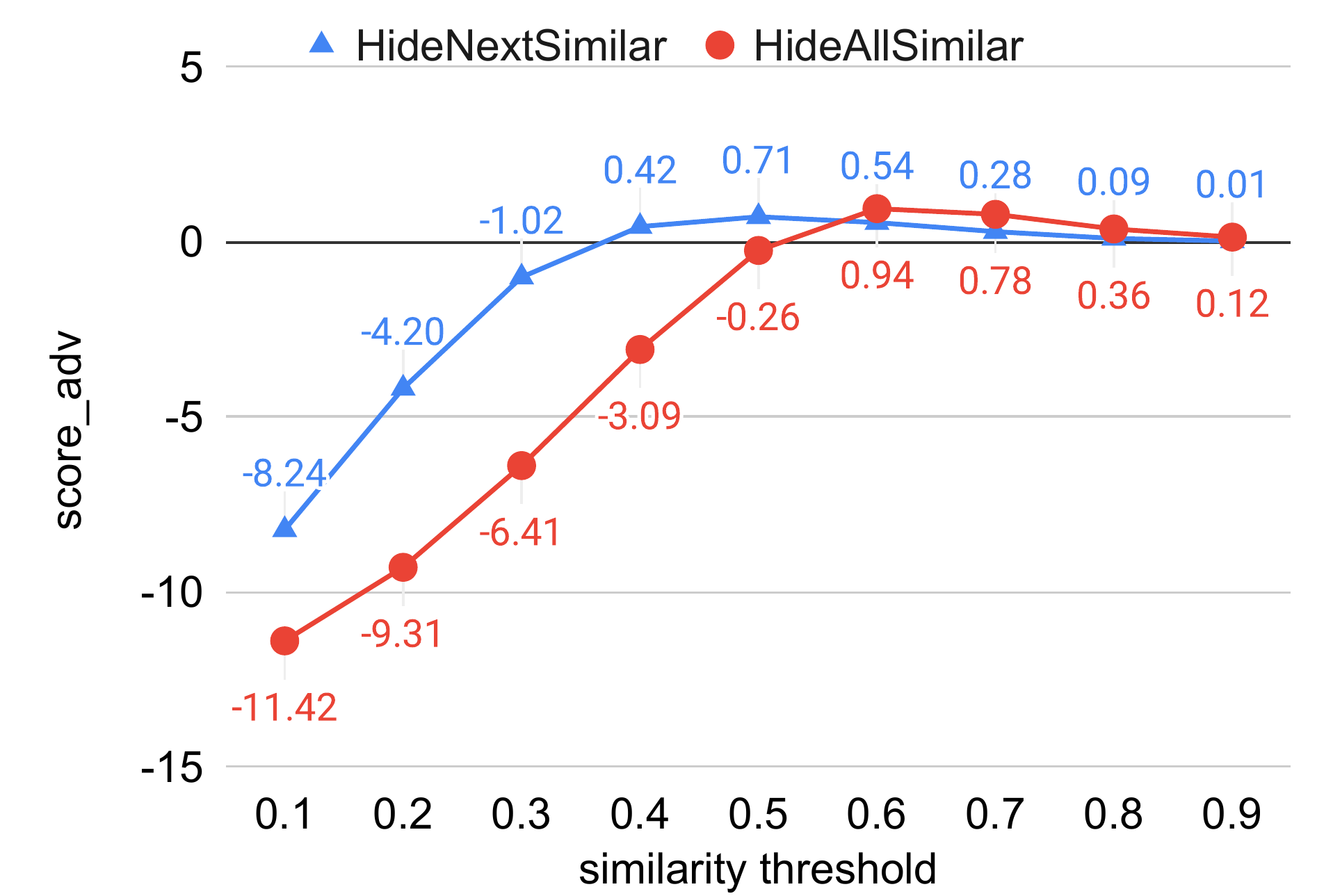}
	\caption{The performance for \textsc{HideNextSimilar} and \textsc{HideAllSimilar} for a range of similarity thresholds. When the threshold is high, it means that only the most similar sentences are hidden.}
	\label{fig:hide_similar_results}
\end{figure}

The related models of \textsc{HideNextSimilar} and \textsc{HideAllSimilar}, which simply hide sentences similar to ones the user swipes away, work surprisingly well, in both moderate and low noise. 
In Figure~\ref{fig:hide_similar_results}, we can see that their performance peaks when the similarity threshold is around 0.5 to 0.6.

\subsubsection{Adapted Generic Summarizers}

We use the following extractive summarizers: LexRank \cite{erkan2004lexrank}, SumBasic \cite{nenkova2005impact}, and TextRank \cite{mihalcea2004textrank}\footnote{Implementations provided by Sumy library, available at \url{https://pypi.python.org/pypi/sumy}.}.

We find that the generic summarizer-based models always perform worse than the control when showing a fixed fraction of the article (\textsc{GenFixed}). The best model of this type used the SumBasic summarizer, showing 75\% of sentences.
%(while more sentences would perform better, performance does not exceed that of the control). 
When dynamically estimating target summary length (\textsc{GenDynamic}), the control is outperformed by only 0.09 points. This is achieved by the SumBasic summarizers with $\epsilon = 0.5$. For both variations, we find that the best hyperparameters tend to be those that make them show the most sentences. 
%Detailed results can be found in the appendix.

\subsubsection{Preference-learning Models}

% \usepackage{booktabs}
% \usepackage{graphicx}
\begin{table}[]
\centering
\resizebox{0.75\columnwidth}{!}{%
\begin{tabular}{@{}rrlrr@{}}
\toprule
\multicolumn{2}{l}{LR (constant $\epsilon$)}                    &  & \multicolumn{2}{l}{LR (decreasing $\epsilon$)}                \\ \midrule
\multicolumn{1}{l}{$\epsilon$} & \multicolumn{1}{l}{$score_{adv}$} &  & \multicolumn{1}{l}{$\beta$} & \multicolumn{1}{l}{$score_{adv}$} \\ \midrule
0   & -7.27         &  & 0.25                 & 0.05                 \\
0.1 & -1.58         &  & 0.5                  & 0.09                \\
0.2 & -0.18         &  & 1                    & 0.15       \\
0.3 & 0.25          &  & 2                    & 0.07                \\
0.4 & 0.34 &  & 4                    & -0.61               \\
0.5 & 0.34          &  & \multicolumn{1}{l}{} & \multicolumn{1}{l}{} \\ \bottomrule
\end{tabular}%
}
\caption{Results for the two \textsc{LR} model version. For the constant-$\epsilon$ variation, a greater $\epsilon$ indicates greater exploration. For the decreasing-$\epsilon$ variation, larger $\beta$ indicates a faster decay in exploration probability. 
%When $\beta=1$, it means that if we are 50\% of the way through the article, $\epsilon=0.5$.
}
\label{tab:lr_results}
\end{table}

%Results for the \textsc{LR} models are in Table~\ref{tab:lr_results}, and results for \textsc{CoverageOpt} are in Table~\ref{tab:coverage_results}.

The \textsc{LR} models out-perform the control, as shown in Table~\ref{tab:lr_results}, but fail to match the simpler approaches. Using a decaying $\epsilon$ actually hurt performance, suggesting that the model is simply not able to learn user preferences fast enough. However, there is a sweet spot for the rate of $\epsilon$ decay at $\beta = 1$.

%We find that \textsc{CoverageOpt}, where the notions of concept weighting, coverage, and target length are built-in, is able to outperform \textsc{LR}. 
We find that \textsc{CoverageOpt} consistently improves with larger initial concept weights ($c$) and a slower concept weight-saturation rate ($\beta$), with the performance plateauing around $\beta = 4$ and $c = 5$.
When both $c$ and $\beta$ are both large, there is a longer exploration phase with more evidence required to indicate that any given concept should be hidden.  
%Detailed results can be found in the appendix.
%These observations suggest that optimistic estimates and a longer exploration phase are beneficial.
%initial preference for showing all sentences (large $c$)... a longer exploration phase (large $\beta$).

\subsection{Efficiency}

\paragraph{Acceptance rate} When measuring the fraction of shown sentences that are accepted, we find no consistent connection to their performance.
For example, the control and the best \textsc{HideNext}, \textsc{HideNextSimilar}, \textsc{HideAllSimilar}, and \textsc{CoverageOpt} models all have rates between 64-66\% in the noisy feedback case.
\textsc{OracleSorted} has the highest however, at 79\%, while \textsc{OracleGreedy} is only at 69\% acceptance. As discussed in Section~\ref{sec:results_privileged}, this is because the sentence set which maximizes the score is not necessarily the same as the set with the highest importance sum.

\paragraph{Speed} The approaches presented here are able to update the summary in real-time. Running on a consumer-grade laptop, each full user-article simulation (which consists of many interactions) takes between 100ms for the slowest model (\textsc{GenFixed} with TextRank), to 2.8ms for \textsc{HideAllSimilar}, to 1.3ms for \textsc{HideNext}.

\subsection{Human Evaluation}

Finally, we run a human evaluation to test a variety of approaches on multiple measures. 
%The results are summarized in Figure~\ref{fig:human_eval}.

% TODO: can add some of the setup information to the appendix

\paragraph{Setup} We selected 10 news articles from a variety of sources and on a variety of topics (such as politics, sports, and science), with an average sentence length of 20.6, and asked 13 volunteers to read articles with the help of randomly assigned \taskname models. In total, we collected 70 trials. Participants were shown sentences one at a time and provided feedback to either accept or reject sentences. They were also able to stop reading each article at any time. 
After reading each article, they were asked several questions about the experience, including the coherence of what they read (how well-connected consecutive sentences were, from 1 to 5) and how easy it was to decide whether to accept or reject sentences (from 1 to 5). We also showed them any unread sentences afterwards in order to determine how many would-be accepted sentences were not shown. Coverage, roughly corresponding to our automated evaluation metric, can then be estimated with the fraction of interesting sentences that were actually shown.

\begin{figure}[ht]%bp]
	\centering
	\includegraphics[width=1\columnwidth]{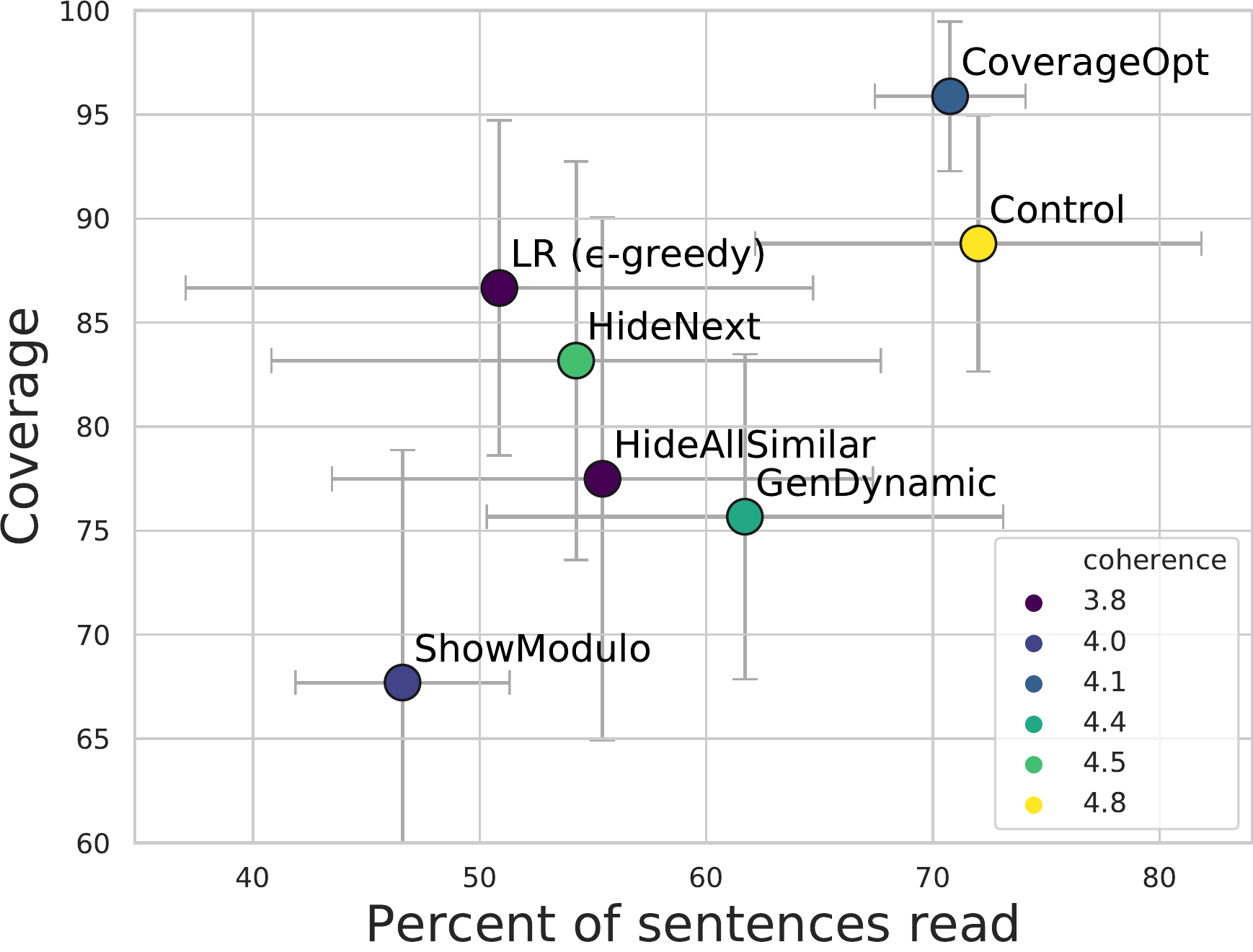}
	\caption{Summary of human evaluation results. Error bars indicate 90\% confidence intervals.}
	\label{fig:human_eval}
\end{figure}

\paragraph{Results} From the human evaluation, we find that making the decision to accept or reject sentences is quite easy, with an average decision-ease rating of 4.4/5.
%However, departing from the assumptions of our user model, people only chose to stop reading the article early 51\% of the time. 
However, departing from the assumptions of our user model, people ended up reading more than an average of 50\% of the articles (up to 70\% for the control). This could influence the relative performance of the various models, with a skew towards models that tend to hide fewer sentences.
%This may be a result of the relatively short lengths of the articles chosen and could influence the relative performance of the various models.
%
We find the acceptance rate to vary from 47\% for \textsc{LR} to 75\% for \textsc{CoverageOpt}, with the remainder around 60\%.
From Figure~\ref{fig:human_eval} we can see that the best model (highest coverage) appears to be \textsc{CoverageOpt}. This is followed by the control and \textsc{LR} model, with their 90\% confidence intervals overlapping. This highlights that achieving good coverage of interesting sentences is not the same as achieving a high acceptance rate. The worst performing model according to both human and automated evaluation is \textsc{ShowModulo}. The remaining four models significantly overlap in their confidence intervals. However, it is interesting to note that \textsc{HideAllSimilar} performs poorer than we would expect. Given the positive correlation between the percent of the article users end up reading and the model coverage, we can guess that this is a result of the model automatically hiding too many sentences. This also leads to low reported summary coherence, as many sentences are skipped. In contrast, the control achieves the highest coherence (since nothing is skipped), with \textsc{CoverageOpt} near the middle of the pack.
%\textsc{HideNext}, which only hides immediately following sentences also achieves moderately high coherence as we might expect.

\section{Conclusion}
\label{sec:conclusion}

In this paper we proposed a new interactive summarization task where the document is automatically refined during the normal flow of reading. By not requiring an explicit query or relying on time-consuming and invasive feedback, relevant information can be conveniently provided for a wide range of user preferences.
We provided an approximate user model and suitable evaluation metric for this task, building upon recent advances in unsupervised summary evaluation.
To guide examination of this new task, we proposed a variety of approaches, and perform both automated and human evaluation.
%ranging from simple heuristics to adapted generic summarizers, to user preference-learning approaches.
%We found that outperforming the control approach of no summarization can be done, by a small but significant margin.
%In particular, we found the simply hiding sentences similar to those disliked by the user worked well. With moderate noise in user feedback, it outperforms an approach which explicitly models user preferences.
Future research on this task includes adapting the interaction model to implicit feedback and trying more advanced approaches.
% running human model preference experiments

\section{Ethical Considerations}

\paragraph{Diversity of viewpoints} The \taskname task is intended for the design of future user-facing applications. By design, these applications have the ability to control what a user reads from a given article. 
It is possible that, when deployed without sufficient care, these tools could exacerbate the ``echo chamber'' effect already produced by automated news feeds, search results, and online communities \citep{pariser2011filter}. 
%If a user starts reading an article and indicates disinterest in text that does not agree with their personal views, they may be even less likely to be exposed to such views through the rest of the article.
%
However, the ability to influence what readers are exposed to can also be leveraged to \textit{mitigate} the echo chamber effect. Rather than considering only what user interests appear to be at a given moment, future \taskname models could incorporate a diversity factor to explicitly encourage exposure to alternative views when possible. The weighting of this factor could be tuned to provide both an engaging reading experience and exposure to a diversity of ideas.

\paragraph{Beneficiaries} As mentioned in Section~\ref{sec:intro}, those most likely to benefit from \taskname applications once successfully deployed will be those using them to read (by saving time and increased engagement) as well as any content publishers who encourage their use.

\iffalse
\section*{Acknowledgments}

The acknowledgments should go immediately before the references. Do not number the acknowledgments section.
\textbf{Do not include this section when submitting your paper for review.}
\fi

\bibliographystyle{acl_natbib}
\bibliography{anthology,acl2021}

\begin{thebibliography}{38}
\expandafter\ifx\csname natexlab\endcsname\relax\def\natexlab#1{#1}\fi

\bibitem[{Avinesh et~al.(2018)Avinesh, Binnig, H{\"a}ttasch, Meyer, and
  {\"O}zyurt}]{avinesh2018sherlock}
PVS Avinesh, Carsten Binnig, Benjamin H{\"a}ttasch, Christian~M Meyer, and
  Orkan {\"O}zyurt. 2018.
\newblock Sherlock: A system for interactive summarization of large text
  collections.
\newblock \emph{Proc. VLDB Endow.}, 11(12):1902--1905.

\bibitem[{Avinesh and Meyer(2017)}]{avinesh2017joint}
PVS Avinesh and Christian~M Meyer. 2017.
\newblock Joint optimization of user-desired content in multi-document
  summaries by learning from user feedback.
\newblock In \emph{Proceedings of the 55th Annual Meeting of the Association
  for Computational Linguistics (Volume 1: Long Papers)}, pages 1353--1363.

\bibitem[{Barratt and Sharma(2018)}]{barratt2018note}
Shane Barratt and Rishi Sharma. 2018.
\newblock \href {http://arxiv.org/abs/1801.01973} {A note on the inception
  score}.
\newblock \emph{arXiv preprint arXiv:1801.01973}.

\bibitem[{B{\"o}hm et~al.(2019)B{\"o}hm, Gao, Meyer, Shapira, Dagan, and
  Gurevych}]{bohm2019better}
Florian B{\"o}hm, Yang Gao, Christian~M Meyer, Ori Shapira, Ido Dagan, and
  Iryna Gurevych. 2019.
\newblock \href {http://arxiv.org/abs/1909.01214} {Better rewards yield better
  summaries: Learning to summarise without references}.
\newblock \emph{arXiv preprint arXiv:1909.01214}.

\bibitem[{Calder et~al.(2009)Calder, Malthouse, and
  Schaedel}]{calder2009experimental}
Bobby~J Calder, Edward~C Malthouse, and Ute Schaedel. 2009.
\newblock An experimental study of the relationship between online engagement
  and advertising effectiveness.
\newblock \emph{Journal of interactive marketing}, 23(4):321--331.

\bibitem[{Conroy et~al.(2005)Conroy, Schlesinger, and
  Stewart}]{conroy2005classy}
John~M Conroy, Judith~D Schlesinger, and Jade~Goldstein Stewart. 2005.
\newblock Classy query-based multi-document summarization.
\newblock In \emph{Proceedings of the 2005 Document Understanding Workshop,
  Boston}. Citeseer.

\bibitem[{Damova and Koychev(2010)}]{damova2010query}
Mariana Damova and Ivan Koychev. 2010.
\newblock Query-based summarization: A survey.

\bibitem[{Dang(2005)}]{dang2005overview}
Hoa~Trang Dang. 2005.
\newblock Overview of duc 2005.
\newblock In \emph{Proceedings of the document understanding conference},
  volume 2005, pages 1--12.

\bibitem[{David and Cambre(2016)}]{david2016screened}
Gaby David and Carolina Cambre. 2016.
\newblock Screened intimacies: Tinder and the swipe logic.
\newblock \emph{Social media+ society}, 2(2):2056305116641976.

\bibitem[{Erkan and Radev(2004)}]{erkan2004lexrank}
G{\"u}nes Erkan and Dragomir~R Radev. 2004.
\newblock Lexrank: Graph-based lexical centrality as salience in text
  summarization.
\newblock \emph{Journal of Artificial Intelligence Research}, 22:457--479.

\bibitem[{Fabbri et~al.(2019)Fabbri, Li, She, Li, and
  Radev}]{alex2019multinews}
Alexander~R. Fabbri, Irene Li, Tianwei She, Suyi Li, and Dragomir~R. Radev.
  2019.
\newblock \href {http://arxiv.org/abs/1906.01749} {Multi-news: a large-scale
  multi-document summarization dataset and abstractive hierarchical model}.

\bibitem[{Gao et~al.(2019)Gao, Meyer, and Gurevych}]{gao2019preference}
Yang Gao, Christian~M Meyer, and Iryna Gurevych. 2019.
\newblock Preference-based interactive multi-document summarisation.
\newblock \emph{Information Retrieval Journal}, pages 1--31.

\bibitem[{Gao et~al.(2020)Gao, Zhao, and Eger}]{gao2020supert}
Yang Gao, Wei Zhao, and Steffen Eger. 2020.
\newblock \href {http://arxiv.org/abs/2005.03724} {Supert: Towards new
  frontiers in unsupervised evaluation metrics for multi-document
  summarization}.
\newblock \emph{arXiv preprint arXiv:2005.03724}.

\bibitem[{Hasselqvist et~al.(2017)Hasselqvist, Helmertz, and
  Kågebäck}]{hasselqvist2017querybased}
Johan Hasselqvist, Niklas Helmertz, and Mikael Kågebäck. 2017.
\newblock \href {http://arxiv.org/abs/1712.06100} {Query-based abstractive
  summarization using neural networks}.

\bibitem[{Hermann et~al.(2015)Hermann, Kocisky, Grefenstette, Espeholt, Kay,
  Suleyman, and Blunsom}]{hermann2015teaching}
Karl~Moritz Hermann, Tomas Kocisky, Edward Grefenstette, Lasse Espeholt, Will
  Kay, Mustafa Suleyman, and Phil Blunsom. 2015.
\newblock Teaching machines to read and comprehend.
\newblock In \emph{Advances in Neural Information Processing Systems}, pages
  1693--1701.

\bibitem[{Hoa(2006)}]{hoa2006overview}
TD~Hoa. 2006.
\newblock Overview of duc 2006.
\newblock In \emph{Document Understanding Conference}.

\bibitem[{Jayarathna and Shipman(2017)}]{jayarathna2017analysis}
Sampath Jayarathna and Frank Shipman. 2017.
\newblock Analysis and modeling of unified user interest.
\newblock In \emph{2017 IEEE International Conference on Information Reuse and
  Integration (IRI)}, pages 298--307. IEEE.

\bibitem[{Kusner et~al.(2015)Kusner, Sun, Kolkin, and
  Weinberger}]{kusner2015word}
Matt Kusner, Yu~Sun, Nicholas Kolkin, and Kilian Weinberger. 2015.
\newblock From word embeddings to document distances.
\newblock In \emph{International conference on machine learning}, pages
  957--966.

\bibitem[{Lin and Och(2004)}]{lin2004automatic}
Chin-Yew Lin and Franz~Josef Och. 2004.
\newblock Automatic evaluation of machine translation quality using longest
  common subsequence and skip-bigram statistics.
\newblock In \emph{Proceedings of the 42nd Annual Meeting of the Association
  for Computational Linguistics (ACL-04)}, pages 605--612.

\bibitem[{Louis and Nenkova(2009)}]{louis2009automatically}
Annie Louis and Ani Nenkova. 2009.
\newblock Automatically evaluating content selection in summarization without
  human models.
\newblock In \emph{Proceedings of the 2009 Conference on Empirical Methods in
  Natural Language Processing}, pages 306--314.

\bibitem[{Louis and Nenkova(2013)}]{louis2013automatically}
Annie Louis and Ani Nenkova. 2013.
\newblock Automatically assessing machine summary content without a gold
  standard.
\newblock \emph{Computational Linguistics}, 39(2):267--300.

\bibitem[{Mihalcea and Tarau(2004)}]{mihalcea2004textrank}
Rada Mihalcea and Paul Tarau. 2004.
\newblock Textrank: Bringing order into text.
\newblock In \emph{Proceedings of the 2004 Conference on Empirical Methods in
  Natural Language Processing}, pages 404--411.

\bibitem[{Mohamed and Rajasekaran(2006)}]{mohamed2006improving}
Ahmed~A Mohamed and Sanguthevar Rajasekaran. 2006.
\newblock Improving query-based summarization using document graphs.
\newblock In \emph{2006 IEEE international symposium on signal processing and
  information technology}, pages 408--410. IEEE.

\bibitem[{Narayan et~al.(2018)Narayan, Cohen, and Lapata}]{Narayan2018DontGM}
Shashi Narayan, Shay~B. Cohen, and Mirella Lapata. 2018.
\newblock \href {http://arxiv.org/abs/1808.08745} {Don't give me the details,
  just the summary! topic-aware convolutional neural networks for extreme
  summarization}.
\newblock \emph{ArXiv}, abs/1808.08745.

\bibitem[{Nenkova and McKeown(2011)}]{nenkova2011automatic}
Ani Nenkova and Kathleen McKeown. 2011.
\newblock \emph{Automatic summarization}.
\newblock Now Publishers Inc.

\bibitem[{Nenkova and Vanderwende(2005)}]{nenkova2005impact}
Ani Nenkova and Lucy Vanderwende. 2005.
\newblock The impact of frequency on summarization.
\newblock \emph{Microsoft Research, Redmond, Washington, Tech. Rep.
  MSR-TR-2005}, 101.

\bibitem[{Pariser(2011)}]{pariser2011filter}
Eli Pariser. 2011.
\newblock \emph{The filter bubble: What the Internet is hiding from you}.
\newblock Penguin UK.

\bibitem[{Peters et~al.(2018)Peters, Neumann, Iyyer, Gardner, Clark, Lee, and
  Zettlemoyer}]{peters2018deep}
Matthew~E Peters, Mark Neumann, Mohit Iyyer, Matt Gardner, Christopher Clark,
  Kenton Lee, and Luke Zettlemoyer. 2018.
\newblock \href {http://arxiv.org/abs/1802.05365} {Deep contextualized word
  representations}.
\newblock \emph{arXiv preprint arXiv:1802.05365}.

\bibitem[{Reimers and Gurevych(2019)}]{reimers2019sentence}
Nils Reimers and Iryna Gurevych. 2019.
\newblock \href {http://arxiv.org/abs/1908.10084} {Sentence-bert: Sentence
  embeddings using siamese bert-networks}.
\newblock \emph{arXiv preprint arXiv:1908.10084}.

\bibitem[{Simpson et~al.(2019)Simpson, Gao, and
  Gurevych}]{simpson2019interactive}
Edwin Simpson, Yang Gao, and Iryna Gurevych. 2019.
\newblock \href {http://arxiv.org/abs/1911.10183} {Interactive text ranking
  with bayesian optimisation: A case study on community qa and summarisation}.
\newblock \emph{arXiv preprint arXiv:1911.10183}.

\bibitem[{Stiennon et~al.(2020)Stiennon, Ouyang, Wu, Ziegler, Lowe, Voss,
  Radford, Amodei, and Christiano}]{stiennon2020learning}
Nisan Stiennon, Long Ouyang, Jeff Wu, Daniel~M Ziegler, Ryan Lowe, Chelsea
  Voss, Alec Radford, Dario Amodei, and Paul Christiano. 2020.
\newblock \href {http://arxiv.org/abs/2009.01325} {Learning to summarize from
  human feedback}.
\newblock \emph{arXiv preprint arXiv:2009.01325}.

\bibitem[{Sun and Nenkova(2019)}]{sun2019feasibility}
Simeng Sun and Ani Nenkova. 2019.
\newblock The feasibility of embedding based automatic evaluation for single
  document summarization.
\newblock In \emph{Proceedings of the 2019 Conference on Empirical Methods in
  Natural Language Processing and the 9th International Joint Conference on
  Natural Language Processing (EMNLP-IJCNLP)}, pages 1216--1221.

\bibitem[{Teevan et~al.(2005)Teevan, Dumais, and
  Horvitz}]{teevan2005personalizing}
Jaime Teevan, Susan~T Dumais, and Eric Horvitz. 2005.
\newblock Personalizing search via automated analysis of interests and
  activities.
\newblock In \emph{Proceedings of the 28th annual international ACM SIGIR
  conference on Research and development in information retrieval}, pages
  449--456.

\bibitem[{Van~Lierde and Chow(2019)}]{van2019query}
Hadrien Van~Lierde and Tommy~WS Chow. 2019.
\newblock Query-oriented text summarization based on hypergraph transversals.
\newblock \emph{Information Processing \& Management}, 56(4):1317--1338.

\bibitem[{Viappiani and Boutilier(2010)}]{viappiani2010optimal}
Paolo Viappiani and Craig Boutilier. 2010.
\newblock Optimal bayesian recommendation sets and myopically optimal choice
  query sets.
\newblock In \emph{Advances in neural information processing systems}, pages
  2352--2360.

\bibitem[{Wu et~al.(2019)Wu, Wu, An, Qi, Huang, Huang, and Xie}]{wu2019neural}
Chuhan Wu, Fangzhao Wu, Mingxiao An, Tao Qi, Jianqiang Huang, Yongfeng Huang,
  and Xing Xie. 2019.
\newblock Neural news recommendation with heterogeneous user behavior.
\newblock In \emph{Proceedings of the 2019 Conference on Empirical Methods in
  Natural Language Processing and the 9th International Joint Conference on
  Natural Language Processing (EMNLP-IJCNLP)}, pages 4876--4885.

\bibitem[{Xu and Lapata(2020)}]{xu2020query}
Yumo Xu and Mirella Lapata. 2020.
\newblock \href {http://arxiv.org/abs/2004.03027} {Query focused multi-document
  summarization with distant supervision}.
\newblock \emph{arXiv preprint arXiv:2004.03027}.

\bibitem[{Yan et~al.(2011)Yan, Nie, and Li}]{yan2011summarize}
Rui Yan, Jian-Yun Nie, and Xiaoming Li. 2011.
\newblock Summarize what you are interested in: An optimization framework for
  interactive personalized summarization.
\newblock In \emph{Proceedings of the 2011 conference on empirical methods in
  natural language processing}, pages 1342--1351.

\end{thebibliography}

\appendix
%\begin{appendices}
\counterwithin{table}{section}
\section{Experimental Setup}

\textbf{Computing infrastructure} All experiments were performed on a machine with an Intel Core i7-6700HQ CPU with 16G RAM and a GeForce GTX 960M GPU.

\textbf{Hyperparameter searches} For parameterized models, grid searches over the following ranges were performed:

\begin{itemize}
    \item \textsc{ShowModulo}: $k \in \{2, 3, 4, 5\}$
    \item \textsc{HideNext}: $n \in \{1, 2, 3, 4\}$
    \item \textsc{HideNextSimilar} and \textsc{HideAllSimilar}: $threshold \in \{0.1, 0.2, 0.3, 0.4, 0.5, 0.6, 0.7, 0.8, 0.9\}$
    \item \textsc{GenFixed}: $frac \in \{0.25, 0.5, 0.75\}$
    \item \textsc{GenDynamic}: $\epsilon \in \{0, 0.1, 0.2, 0.3, 0.4, 0.5\}$ 
    \item \textsc{LR} (constant $\epsilon$): $\epsilon \in \{0, 0.1, 0.2, 0.3, 0.4, 0.5\}$ 
    \item \textsc{LR} (decreasing $\epsilon$): $\beta \in \{0.25, 0.5, 1, 2, 4\}$ 
    \item \textsc{CoverageOpt}: $\beta \in \{0.25, 0.5, 1, 2, 4\}$ and $c \in \{0, 1, 2, 3, 4\}$
\end{itemize}

\section{Detailed Results}

Detailed results for those models without full results reported in the paper are shown here. For \textsc{ShowModulo} and \textsc{HideNext}, results are shown in Table~\ref{tab:simplest_results}. For summarizer-based models, results are shown in Table~\ref{tab:summarizer_results}. For \textsc{CoverageOpt}, results are shown in Table~\ref{tab:coverage_results}.

% \usepackage{booktabs}
% \usepackage{graphicx}
\begin{table}[]
\centering
\resizebox{0.65\columnwidth}{!}{%
\begin{tabular}{@{}rrrrr@{}}
\toprule
\multicolumn{2}{c}{\textsc{ShowModulo}} &
  \multicolumn{1}{c}{} &
  \multicolumn{2}{c}{\textsc{HideNext}} \\ \midrule
\multicolumn{1}{l}{$k$} &
  \multicolumn{1}{l}{$score_{adv}$} &
  \multicolumn{1}{l}{} &
  \multicolumn{1}{l}{$n$} &
  \multicolumn{1}{l}{$score_{adv}$} \\ \midrule
2 & -3.32  &  & 1 & 0.45 \\
3 & -7.06  &  & 2 & 0.51 \\
4 & -9.87  &  & 3 & 0.19 \\
5 & -12.00 &  & 4 & -0.41 \\ \bottomrule
\end{tabular}%
}
\caption{Results for the first two simple heuristic models. For \textsc{ShowModulo}, every $k^{th}$ sentence is shown. For \textsc{HideNext}, the $n$ sentences following a swiped one are hidden.}
\label{tab:simplest_results}
\end{table}

% \usepackage{booktabs}
% \usepackage{graphicx}
\begin{table}[]
\centering
\resizebox{\columnwidth}{!}{%
\begin{tabular}{@{}lrrrlll@{}}
\toprule
\textbf{}           & \multicolumn{3}{c}{frac (for \textsc{GenFixed})}  &                           &                           &                                   \\ \cmidrule(l){2-7} 
summarizer & 0.25        & 0.5       & 0.75               &  &  &  \\ \midrule
LexRank    & -11.18      & -3.77     & -0.79              &  &  &  \\
SumBasic   & -10.75      & -3.22     & -0.19     &  &  &  \\
TextRank   & -12.28      & -4.99     & -1.53              &  &  &  \\ \midrule
\textbf{}           & \multicolumn{6}{c}{$\epsilon$ (for \textsc{GenDynamic})}                                                                                         \\ \cmidrule(l){2-7} 
summarizer & 0     & 0.1   & 0.2   & \multicolumn{1}{r}{0.3}   & \multicolumn{1}{r}{0.4}   & \multicolumn{1}{r}{0.5}           \\ \midrule
LexRank    & -1.37 & -0.53 & -0.22 & \multicolumn{1}{r}{-0.07} & \multicolumn{1}{r}{0.01}  & \multicolumn{1}{r}{0.06} \\
SumBasic   & -3.19 & -1.47 & -0.72 & \multicolumn{1}{r}{-0.28} & \multicolumn{1}{r}{-0.05} & \multicolumn{1}{r}{0.09} \\
TextRank   & -1.95 & -1.02 & -0.59 & \multicolumn{1}{r}{-0.31} & \multicolumn{1}{r}{-0.18} & \multicolumn{1}{r}{-0.08}         \\ \bottomrule
\end{tabular}%
}
\caption{Results for the two variations of adapted generic summarizer models, for each of three extractive summarizers tested. For \textsc{GenFixed}, $frac$ indicates what fraction of the document is shown, after first sorting sentences by importance. For \textsc{GenDynamic}, $\epsilon$ is used for $\epsilon$-greedy exploration to estimate length preference.
}
\label{tab:summarizer_results}
\end{table}

% \usepackage{booktabs}
% \usepackage{graphicx}
\begin{table}[]
\centering
\resizebox{0.6\columnwidth}{!}{%
\begin{tabular}{@{}rrrrrr@{}}
\toprule
\multicolumn{1}{l}{} & \multicolumn{5}{c}{$\beta$} \\ \cmidrule(l){2-6} 
\multicolumn{1}{l}{$c$} & \textbf{1/4} & \textbf{1/2} & \textbf{1} & \textbf{2} & \textbf{4} \\ \midrule
\textbf{0}           & 0.12  & 0.22 & 0.33 & 0.42 & 0.50 \\
\textbf{1}           & 0.51  & 0.50 & 0.51 & 0.52 & 0.55 \\
\textbf{2}           & 0.49  & 0.57 & 0.60 & 0.59 & 0.59 \\
\textbf{3}           & 0.50  & 0.53 & 0.61 & 0.63 & 0.63 \\
\textbf{4}           & 0.49  & 0.50 & 0.59 & 0.64 & 0.64 \\
\textbf{5}           & 0.49  & 0.50 & 0.55 & 0.64 & \textbf{0.65} \\ \bottomrule
\end{tabular}%
}
\caption{Results for the \textsc{CoverageOpt} model. $c$ controls the initial estimate for concept importances and $\beta$ controls how smoothly a concept shifts between important and unimportant.}
\label{tab:coverage_results}
\end{table}

\section{Human Evaluation}

\begin{figure}[ht]%bp]
	\centering
	\includegraphics[width=0.65\columnwidth]{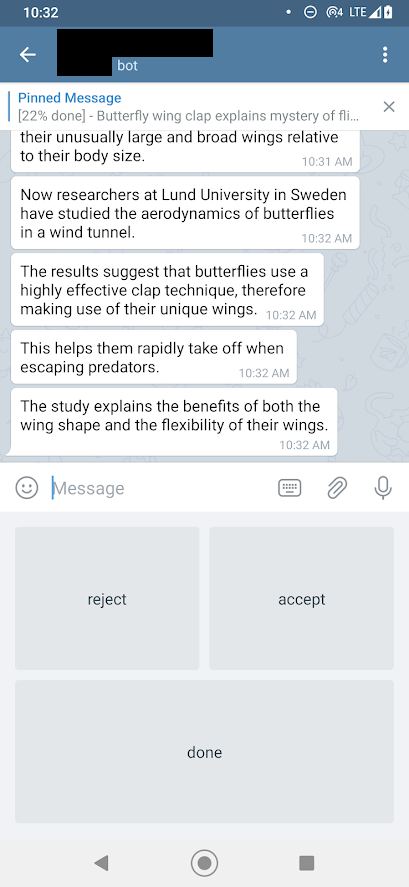}
	\caption{A screenshot of the demo in action. For each sentence, users were able to accept, reject, or stop reading the article at that point.}
	\label{fig:chatbot}
\end{figure}

Human evaluation was performed via a chatbot deployed on the Telegram chat app\footnote{\url{https://telegram.org/}} using their convenient API\footnote{\url{https://core.telegram.org/bots/api}}. A screenshot of the chatbot serving as a simple \taskname interface is shown in Figure~\ref{fig:chatbot}. To participate, volunteers were instructed to engage with the publicly accessible bot in the app and follow instructions provided therein.

%\end{appendices}
\newpage

\end{document}